\documentclass[10pt, a4paper]{article}

\usepackage[final]{lrec2026} 
\usepackage{comment}
\usepackage{booktabs}
\usepackage{tcolorbox}
\tcbuselibrary{listings, skins, breakable}
\usepackage{tabularx}
\usepackage{adjustbox}
\tcbuselibrary{breakable}
\usepackage{graphicx}
\usepackage{caption}
\usepackage{subcaption}
\usepackage{float}
\usepackage{microtype}

\title{Structured Disagreement in Health-Literacy Annotation:\\
Epistemic Stability, Conceptual Difficulty, and Agreement-Stratified Inference}

\name{Olga Kellert$^{*}$\thanks{$^{*}$Main and Corresponding Author.}, Sriya Kondury, Candice Koo, Nemika Tyagi, Steffen Eikenberry}

\address{
Arizona State University \\
\{Olga.Kellert, skondury, ckoo4, ntyagi8, seikenbe\}@asu.edu
}

\abstract{Annotation pipelines in Natural Language Processing (NLP) commonly assume a single latent ground truth per instance and resolve disagreement through label aggregation. Perspectivist approaches challenge this view by treating disagreement as potentially informative rather than erroneous. We present a large-scale analysis of graded health-literacy annotations from 6,323 open-ended COVID-19 responses collected in Ecuador and Peru. Each response was independently labeled by multiple annotators using proportional correctness scores, reflecting the degree to which responses align with normative public-health guidelines, allowing us to analyze the full distribution of judgments rather than aggregated labels. Variance decomposition shows that question-level conceptual difficulty accounts for substantially more variance than annotator identity, indicating that disagreement is structured by the task itself rather than driven by individual raters. Agreement-stratified analyses further reveal that key social-scientific effects, including country, education, and urban–rural differences, vary in magnitude and in some cases reverse direction across levels of inter-annotator agreement. These findings suggest that graded health-literacy evaluation contains both epistemically stable and unstable components, and that aggregating across them can obscure important inferential differences. We therefore argue that strong perspectivist modeling is not only conceptually justified but statistically necessary for valid inference in graded interpretive tasks.
 \\ \newline \Keywords{Perspectivist NLP, Health literacy, Low-resource languages, Annotation disagreement}}
\begin{document}

\maketitleabstract

\section{Introduction}

Manual annotation underlies most NLP systems. The dominant paradigm assumes that each instance has a single latent ground truth and resolves disagreement through aggregation methods such as majority voting or probabilistic label modeling. Within this framework, variability is typically interpreted as annotation noise or insufficient guideline clarity. Perspectivist approaches challenge this assumption by treating disagreement as potentially informative rather than erroneous \cite{basile2021manifesto}. While perspectivism has been widely discussed in overtly subjective tasks such as toxicity or stance detection \cite{davani2022,kanclerz2021}, less attention has been paid to more objective graded interpretive tasks. This raises a broader question: when does disagreement reflect instability in the task itself rather than annotator unreliability? Health-literacy evaluation provides a compelling case. Assessing whether an open-ended response to a public-health question is correct often involves interpreting partial knowledge, implicit reasoning, and degrees of completeness. Such judgments lie between fact verification and subjective stance evaluation, particularly in multilingual and low-resource settings where linguistic variation and unequal access to formal health information widen interpretive space.

In this paper, we analyze structured disagreement in graded COVID-19 health-literacy annotation across Ecuador and Peru. The dataset consists of 6,323 open-ended response-question items annotated independently by four raters using proportional correctness scores on a five-point scale, reflecting the degree to which each response aligns with normative answers derived from WHO and national public health guidelines, yielding 17,305 annotation-level observations. The corpus includes Spanish and Quechua-Kichwa responses along with sociodemographic metadata and represents one of the largest publicly releasable datasets linking graded health-literacy judgments with Indigenous Andean communities. We examine disagreement using variance decomposition and agreement-stratified inference. Question-level conceptual difficulty explains substantially more variance than annotator identity, indicating that disagreement is primarily task-structured rather than rater-driven. 

Although lexical modeling captures substantial signal in responses, it does not eliminate epistemic variability. Crucially, social-scientific effects such as education and urban-rural differences vary in magnitude and, in some cases, direction depending on levels of inter-annotator agreement. Aggregation, therefore, obscures important inferential differences. In this work, we argue that graded health-literacy evaluation contains both epistemically stable and unstable components and that strong perspectivist modeling is statistically necessary for valid inference in graded interpretive tasks.


\section{Background and Related Work}
Perspectivism distinguishes three central concepts \cite{frenda2025survey}:
\begin{itemize}
\item \textbf{Disagreement}: observable variability in labels.
\item \textbf{Subjectivity}: interpretive dependence on individual perspective.
\item \textbf{Reliability}: annotator consistency independent of stance.
\end{itemize}

Disagreement may arise from subjectivity, ambiguity, or conceptual difficulty rather than annotator incompetence \cite{plank2014,uma2022}. Weak perspectivism preserves disaggregated labels, whereas strong perspectivism incorporates disagreement into model training, evaluation, and explanation \cite{basile2021manifesto}. The latter treats variability as a signal rather than noise. Most perspectivist research has focused on clearly subjective domains such as hate speech, stance classification, and aggressiveness detection in user-specific settings \cite{davani2022,kanclerz2021}. However, disagreement also appears in tasks typically framed as objective, including semantic similarity and inference \cite{biester2022}. This suggests that epistemic instability may arise not only from annotator differences but also from properties of the task itself.

Public-health evaluation has traditionally relied on aggregated correctness judgments and closed-ended assessment tools, leaving graded interpretive variability underexplored. Prior work on health literacy assessment \cite{altin2014} and COVID-19 knowledge \cite{meneses2020, mejia2022} further suggests that educational background and community context can shape how public health information is interpreted. In multilingual and Indigenous settings, these challenges are especially pronounced. By examining health-literacy annotations through a perspectivist lens, this study extends disagreement-aware analysis to a graded public-health evaluation domain that combines factual knowledge, conceptual difficulty, and socially structured variation.
\section{Data}

\subsection{Survey Design}

Open-ended COVID-19 knowledge responses were collected in Ecuador and Peru as part of a broader cross-national health communication study \cite{kellert_inpress}. The instrument included questions covering transmission, symptoms, vaccination, risk groups, mask use, and protective measures. The survey was designed to elicit open-ended responses rather than forced-choice answers in order to capture graded health literacy and interpretive variability across respondents.
\paragraph{Size}

\begin{itemize}
\item 17,305 annotation-level observations
\item 6,323 response-question items
\item 25 question identifiers
\item 6,280 non-empty responses
\end{itemize}

\subsection{Dataset \& Participants}

This corpus links graded health-literacy judgments with respondent-level sociodemographic metadata and self-reported information sources in historically underserved Quechua/Kichwa-speaking communities in Peru and Ecuador. It combines (i) open-ended responses, (ii) graded expert annotations, and (iii) metadata on language use, education, and location. The dataset is therefore relevant to public-health research, low-resource NLP, and perspectivist annotation studies that require naturally occurring interpretive variability rather than artificially constructed ambiguity.

\paragraph{Participants.}
Participants (N = 299) were recruited through targeted snowball sampling, a non-probability recruitment method in which referrals are guided toward specific subgroups of interest (e.g., rural, Indigenous, and linguistically diverse communities) in selected urban and rural sites (Lima and Apurímac in Peru; Cañar and El Tambo in Ecuador). The strategy aimed to capture variation in language use, education, and information access, particularly among Indigenous communities. Although non-probabilistic, the sampling was designed to increase representation from rural and linguistically diverse populations central to the study aims.

\paragraph{Survey items and analytic subset.}
The questionnaire included 30 items (including predominantly open-ended questions, along with a small number of structured items and metadata questions). For inferential analyses, we selected seven open-ended questions that were semantically specific enough to allow comparison with official WHO and national public-health guidance and that were frequently answered across participants. These items were selected prior to analysis and were used to construct composite knowledge scores and question-level models. In total, 18 questions were annotated for correctness, of which seven open-ended questions were used for the primary inferential analyses reported in this paper.

\paragraph{Normative references and fieldwork.}
Normative answers were derived from official national health ministry materials and WHO recommendations available at the time of fieldwork and were used to develop the annotation rubric. Data were collected from November to December 2022 by trained local fieldworkers in participants’ preferred language (Spanish, Quechua, or Kichwa). All responses were anonymized prior to analysis.

\subsection{Annotation Protocol}

\paragraph{Annotators.}
Four annotators participated in the labeling process. All were undergraduate students (aged 19-22) at Arizona State University, including two studying data science and two with backgrounds in biology and healthcare. Two annotators were fluent in Spanish and additionally assisted with dataset translation. This combination of technical and health-related expertise was intended to support both analytical consistency and domain-informed evaluation. All annotators were drawn from the same institutional context, which may introduce shared interpretive biases. We mitigate this by focusing on agreement structure and question-level variability rather than treating annotations as independent ground-truth judgments.

\paragraph{Scoring procedure.}
Responses were evaluated using a proportional correctness scoring scheme anchored to a five-point scale:

\[
\{0, 0.25, 0.5, 0.75, 1.0\}
\]

Proportional correctness reflects the degree to which a response aligns with normative answers derived from official national health ministry materials and WHO guidelines. Scoring was adapted to question type. Binary questions were scored as 0 (incorrect) or 1 (correct). For structured selection questions, scores were assigned proportionally based on the number of correct options selected. For open-ended responses, annotators assigned scores based on the extent to which key concepts from the normative answer were present, with partial matches receiving intermediate values.

While the five-point scale served as a common reference, some items admitted finer-grained proportional scores depending on the number of possible correct elements. Label distributions were skewed toward fully correct responses (score = 1), with a substantial proportion of partially correct responses (e.g., score = 0.5), indicating that intermediate values capture meaningful partial knowledge rather than annotator uncertainty.

Annotators worked independently and did not discuss individual cases during scoring. They were blinded to respondent sociodemographic metadata to minimize bias. Prior to annotation, raters received training using a pilot subset of responses to ensure consistent interpretation of the scoring rubric. For baseline lexical modeling, proportional scores were binarized into incorrect (< 0.5) and correct ($\ge 0.5$). This threshold reflects the distinction between predominantly incorrect and predominantly correct responses while preserving graded variation for subsequent disagreement analyses.

\section{Agreement and Baseline Modeling}

\paragraph{Metrics.}
Throughout the analysis, \textit{accuracy} refers to the degree to which a response matches the normative answer derived from official public-health guidance. We report the following three measures throughout the paper:

\begin{itemize}
\item \textbf{Weighted Fleiss' $\kappa$.} We use the weighted version of Fleiss' $\kappa$, which accounts for ordinal distance between categories and therefore captures partial agreement more appropriately than the unweighted variant. Following conventional interpretation guidelines, values below 0.40 indicate low agreement, values between 0.40 and 0.60 moderate agreement, and values above 0.60 substantial agreement \cite{fleiss1971}.

\item \textbf{TF-IDF (Term Frequency-Inverse Document Frequency).} TF-IDF weights tokens by their frequency within a document relative to their frequency across the corpus. This representation captures discriminative lexical patterns and is used here as a baseline feature space for predicting binary correctness labels.

\item \textbf{Intraclass Correlation (ICC).} To quantify the structure of disagreement, we report intraclass correlation coefficients (ICC), which estimate the proportion of total variance attributable to grouping factors (e.g., question or annotator). A high ICC for questions indicates that a substantial portion of the variance is structured by question-level properties rather than annotator identity.
\end{itemize}

Inter-annotator agreement, measured using weighted $\kappa$ across the seven analytic questions, ranged from 0.42 to 0.70, indicating moderate to substantial agreement. As a lexical baseline, we trained a logistic regression classifier over TF-IDF features to predict binary correctness (incorrect $< 0.5$, correct $\geq 0.5$). Model performance was evaluated using stratified 5-fold cross-validation to preserve class balance across splits. The classifier achieved:

\[
\textbf{Accuracy} = 0.8398 \quad (SD = 0.0062)
\]

The reported accuracy reflects the mean performance across held-out folds, with standard deviation indicating fold-level variability. This result demonstrates substantial lexical predictability: responses judged as correct and incorrect exhibit systematic differences in word usage. However, lexical separability does not imply epistemic stability. The classifier captures surface-level signal in token distributions but does not eliminate structured disagreement. Despite this predictive performance, residual variability remains strongly tied to question-level conceptual difficulty and agreement regimes. High lexical predictability therefore does not equate to inferential stability.

\paragraph{Variance Structure.}
To quantify the structure of disagreement, we estimated a mixed-effects model with random intercepts for question and annotator using proportional correctness scores as the dependent variable. Total variance was:

\[
\sigma^2_{total} = 0.13196
\]

\begin{table}[h]
\centering
\begin{tabular}{lrr}
\toprule
Component & Variance & Percent \\
\midrule
Question-level & 0.04451 & 33.73\% \\
Annotator-level & 0.00151 & 1.14\% \\
Residual & 0.08594 & 65.13\% \\
\bottomrule
\end{tabular}
\caption{Variance decomposition.}
\end{table}

The intraclass correlation for questions ($ICC_{question} \approx 0.337$) substantially exceeds that for annotators ($ICC_{annotator} \approx 0.011$), indicating that disagreement is primarily structured by question-level conceptual difficulty rather than annotator inconsistency. In other words, variability reflects properties of the task more than properties of the raters.

\section{Question-Level Variability and Education Gradients}

The variance decomposition results indicated that disagreement clusters at the question level rather than at the annotator level. To further examine the source of this variability, we conducted a per-question variability analysis across all accuracy items.

\subsection{Ranking Questions by Variability}

For each question, we computed the variance of accuracy scores across respondents. The highest-variance questions were:

\begin{itemize}
\item 2.11. Is it possible to reuse masks? (Var = 0.162)
\item 2.13. What is the safe distance between people? (Var = 0.149)
\item 2.5. Asymptomatic infection (Var = 0.104)
\item 2.10. How do masks prevent illness? (Var = 0.103)
\item 2.1. How does COVID-19 spread? (Var = 0.101)
\item 2.14. Who are the risk groups for COVID-19?
\end{itemize}

These items assess transmission mechanisms, mask efficacy, asymptomatic infection, and protective distancing, which are concepts that require both biomedical knowledge and interpretive reasoning.

\subsection{Education Gradients}

Education range refers to the difference in mean accuracy between the lowest and highest education groups on a 0-1 scale. To distinguish conceptual ambiguity from differences in knowledge across education levels, we computed the range in mean accuracy between the lowest and highest education groups for each question. The highest-variance questions exhibited large differences between education groups:

\begin{itemize}
\item 2.1. Spread of COVID-19: education range = 0.515
\item 2.5. Asymptomatic infection: education range = 0.473
\item 2.10. Mask mechanism: education range = 0.455
\item 2.11. Mask reuse: education range = 0.400
\item 2.13. Safe distancing: education range = 0.395
\item 2.14. Risk groups: education range = 0.129
\end{itemize}


\subsection{Interpretation}

High variability in accuracy does not appear to arise solely from annotation ambiguity. Instead, it systematically co-occurs with large education-based differences in performance. Disagreement therefore clusters around conceptually demanding questions where knowledge is unevenly distributed across respondents. This pattern suggests that variability reflects the interaction between task-level conceptual difficulty and socially structured knowledge differences, rather than pure subjectivity or annotator inconsistency. Questions with higher variance are precisely those for which biomedical mechanisms and implicit reasoning are required, making partial understanding more likely and graded judgments more sensitive to interpretive nuance. In particular, the presence of intermediate scores reflects graded partial correctness rather than binary uncertainty, reinforcing that disagreement arises from differing degrees of knowledge rather than purely subjective interpretation.

Moreover, high-variance items tend to exhibit less stable patterns across agreement levels, indicating that education-related effects depend on how clearly and consistently responses can be evaluated. Disagreement in this setting thus reflects structured epistemic difficulty rather than arbitrary labeling variation.




\section{Agreement-Stratified Inference}

To examine whether substantive conclusions vary across levels of epistemic stability, we stratified questions by inter-annotator agreement using Fleiss’ $\kappa$. Items were classified as High ($\kappa > 0.60$), Medium ($0.40 \le \kappa \le 0.60$), and Low ($\kappa < 0.40$). For each stratum, we recomputed sociodemographic effects to assess whether relationships remain stable or shift across agreement levels. Sociodemographic differences were evaluated using standard statistical tests appropriate to each comparison (e.g., t-tests for binary group comparisons and one-way ANOVA for multi-level factors).

\subsection{Country Effects}

\begin{figure}[h]
\centering
\includegraphics[width=7cm]{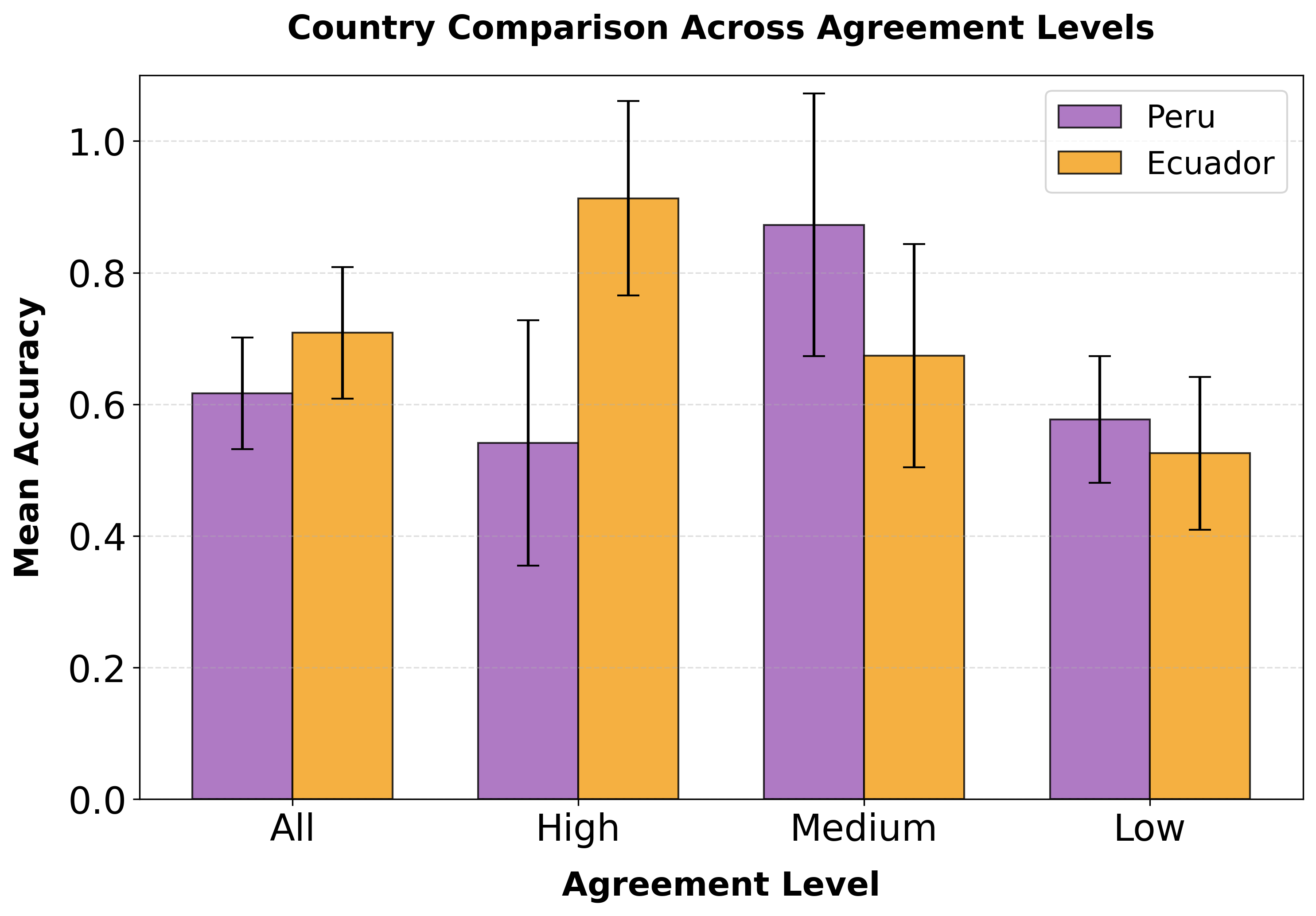}
\caption{Country Comparison Across Agreement Levels
}
\label{fig:country_effects}
\end{figure}

As shown in Figure \ref{fig:country_effects}, agreement levels vary across countries. Across all questions, Ecuador exhibits higher mean accuracy than Peru. Under high agreement, the country difference strengthens substantially. Under medium agreement, however, the direction reverses, with Peru outperforming Ecuador. Under low agreement, the difference attenuates but remains moderate in magnitude. Thus, both the magnitude and direction of country differences vary across agreement levels.

\subsection{Gender Effects}
\begin{figure}[h]
\centering
\includegraphics[width=7cm]
{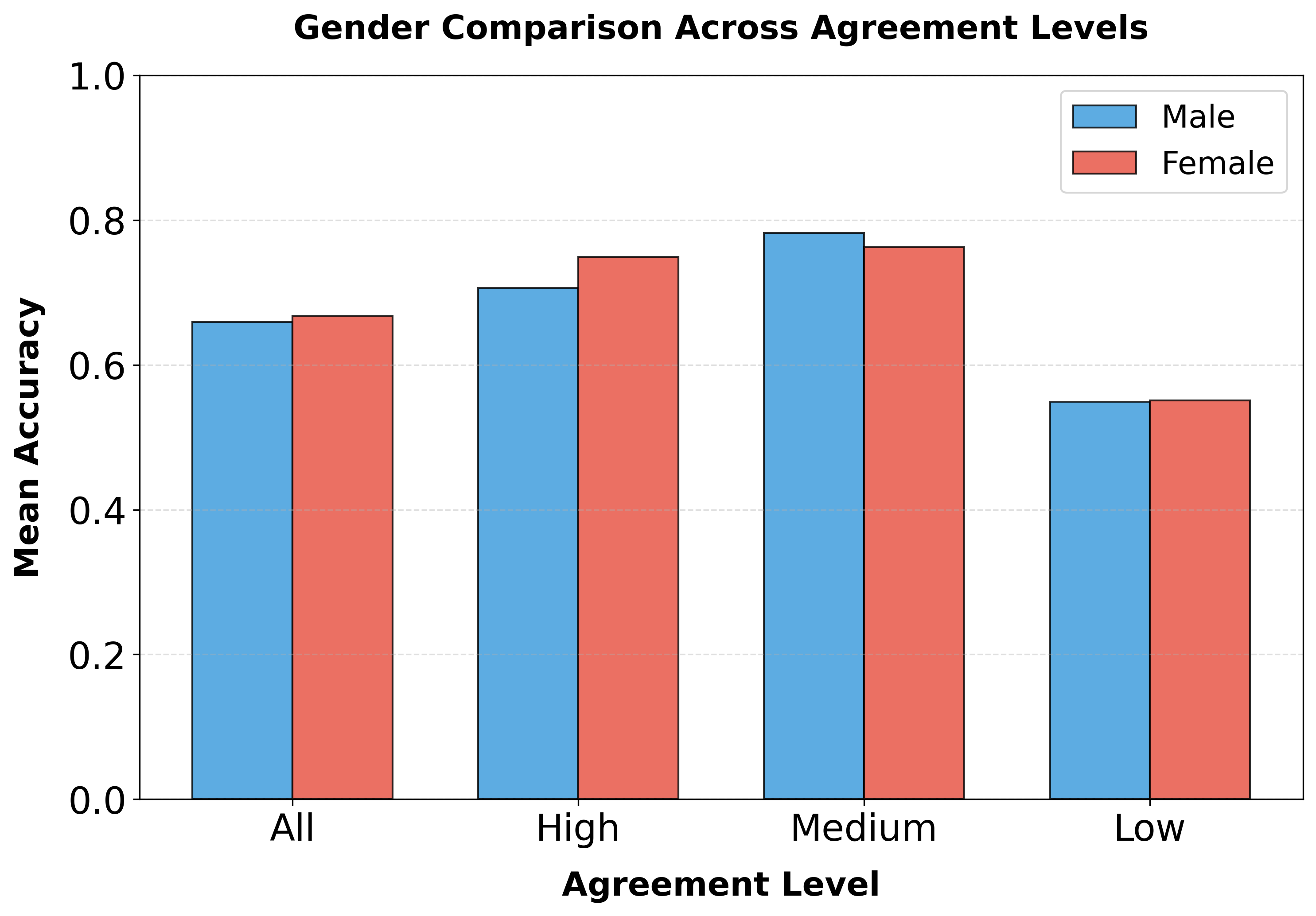}
\caption{ Gender Comparison Across Agreement Levels
}
\label{fig:gen_effects}
\end{figure}

Gender shows no significant differences across any agreement level, as shown in Figure \ref{fig:gen_effects}. The null result remains stable under high, medium, and low agreement, indicating that the absence of gender effects is not driven by aggregation.

\subsection{Education Effects}

\begin{figure}[h]
\centering
\includegraphics[width=7cm]{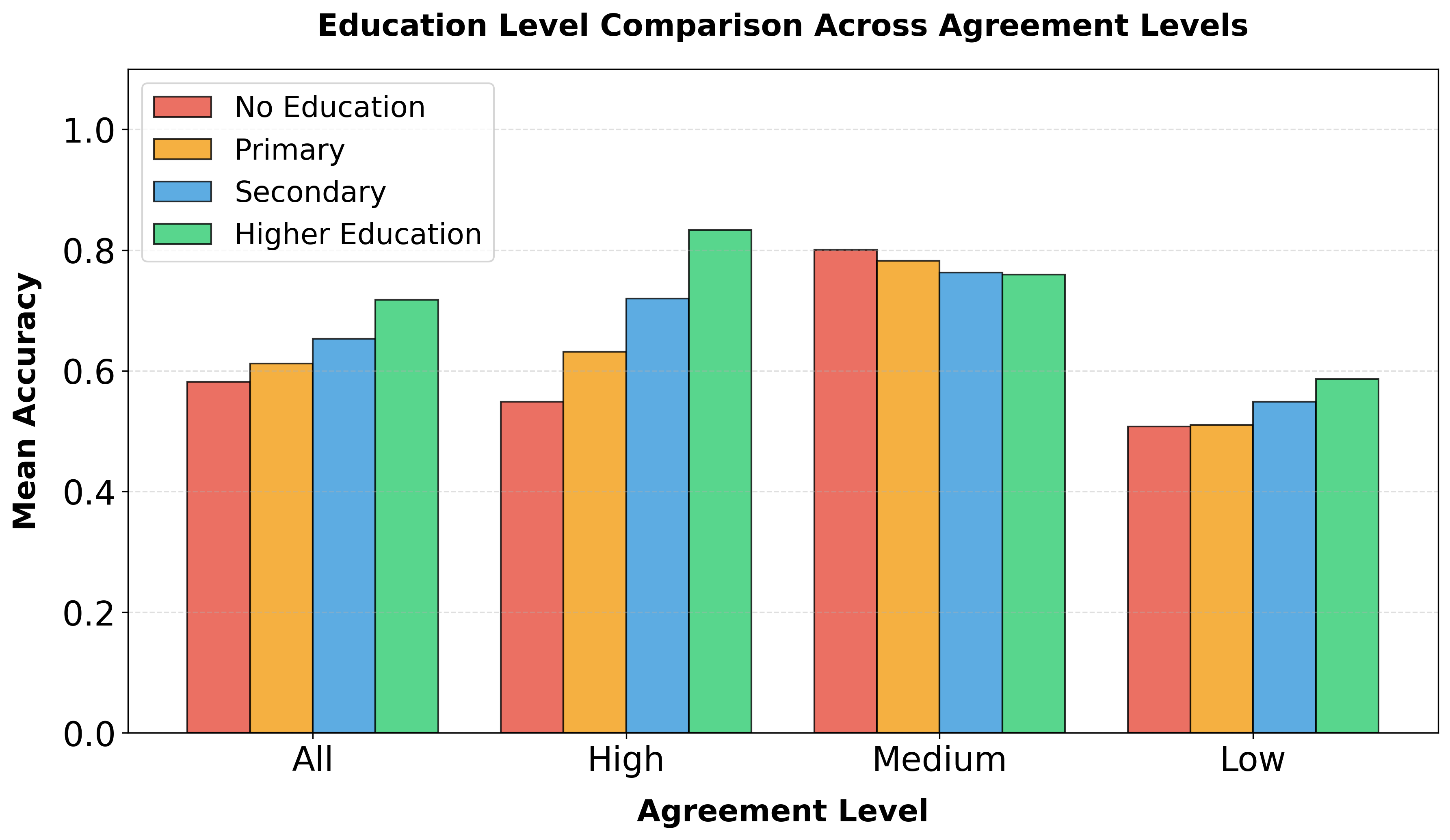}
\caption{ Education Comparison Across Agreement Levels
}
\label{fig:ed_effects}
\end{figure}

Education is a significant predictor in the aggregated model as shown in Figure \ref{fig:ed_effects}. Under high agreement, the education gradient remains strong ($p<.001$). Under medium agreement, however, the effect disappears entirely ($p = .837$), indicating no systematic education-based difference. Under low agreement, the effect re-emerges, though with reduced magnitude.

Thus, education-based inference collapses under moderate disagreement and is highly sensitive to the agreement level.

\subsection{Urban-Rural Effects}

\begin{figure}[h]
\centering
\includegraphics[width=7cm]
{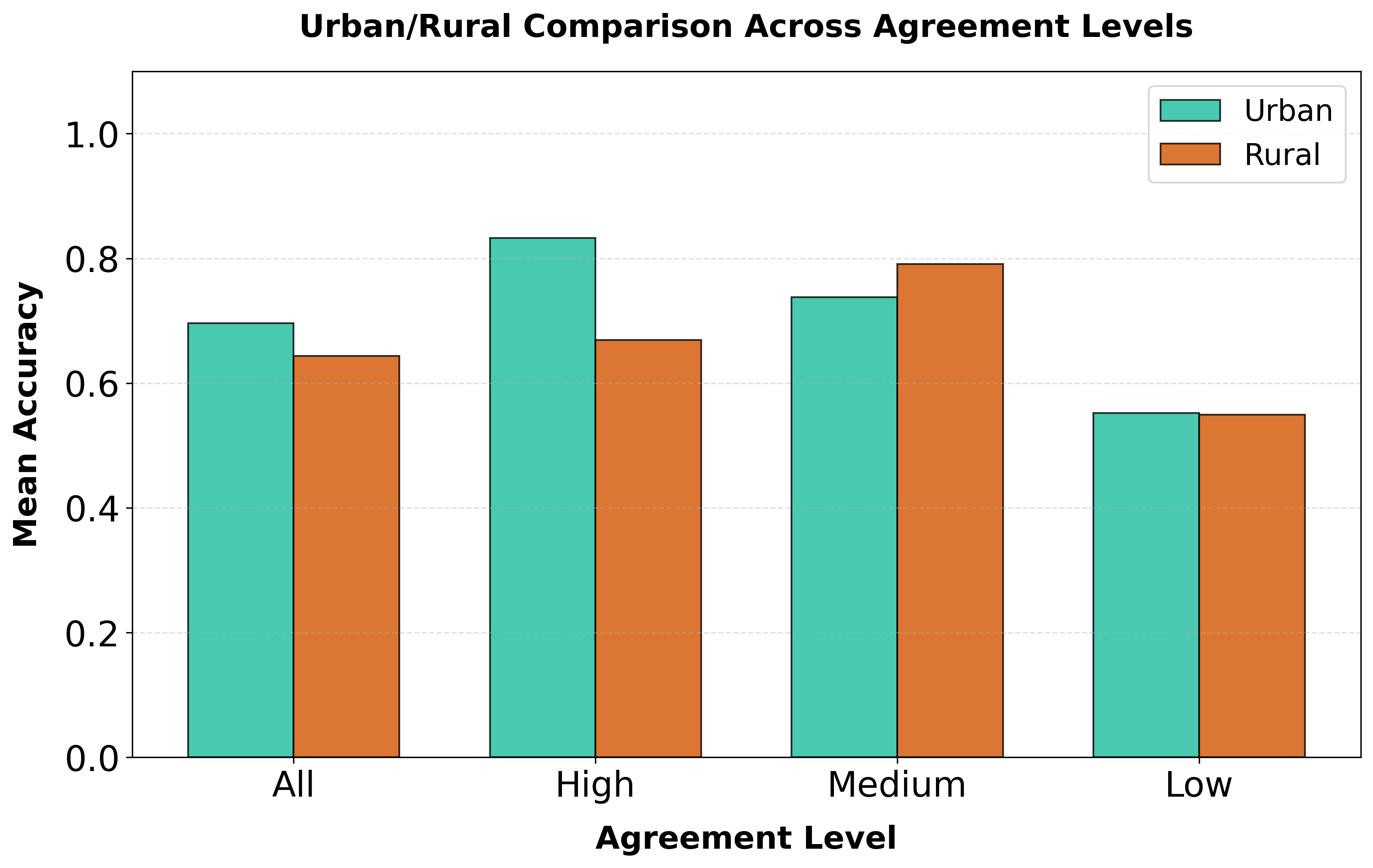}
\caption{ Urban and Rural Comparisons Across Agreement Levels
}
\label{fig:urb_rur_effects}
\end{figure}

As shown in Figure \ref{fig:urb_rur_effects}, urban respondents outperform rural respondents in the aggregated data (Urban: $n=117$, $M=0.6961$; Rural: $n=181$, $M=0.6438$). Under high agreement, the urban advantage strengthens ($t=6.2274$, $p<.001$). Under medium agreement, the effect reverses ($t=-2.12$, $p=.0351$), indicating higher accuracy in rural respondents. Under low agreement, the difference disappears entirely ($t=0.24$, $p=.8124$).

Again, we see that inference is agreement-dependent. Figure 5 synthesizes the agreement-stratified results across predictors, demonstrating that effect magnitudes, and in some cases even their direction, shift across high, medium, and low agreement regimes, thereby exposing the instability introduced by epistemic variation. These reversals under medium agreement suggest that responses in this regime are more ambiguous or inconsistently interpretable, leading to unstable group comparisons. In such cases, annotators may agree on partial correctness but differ in how strongly responses align with normative criteria, which can attenuate or invert observed effects. This highlights that medium-agreement items occupy a transitional zone between clearly interpretable and highly ambiguous responses, where inference is particularly sensitive to how correctness is operationalized.

\begin{figure}[h]
\centering
\includegraphics[width=7cm]{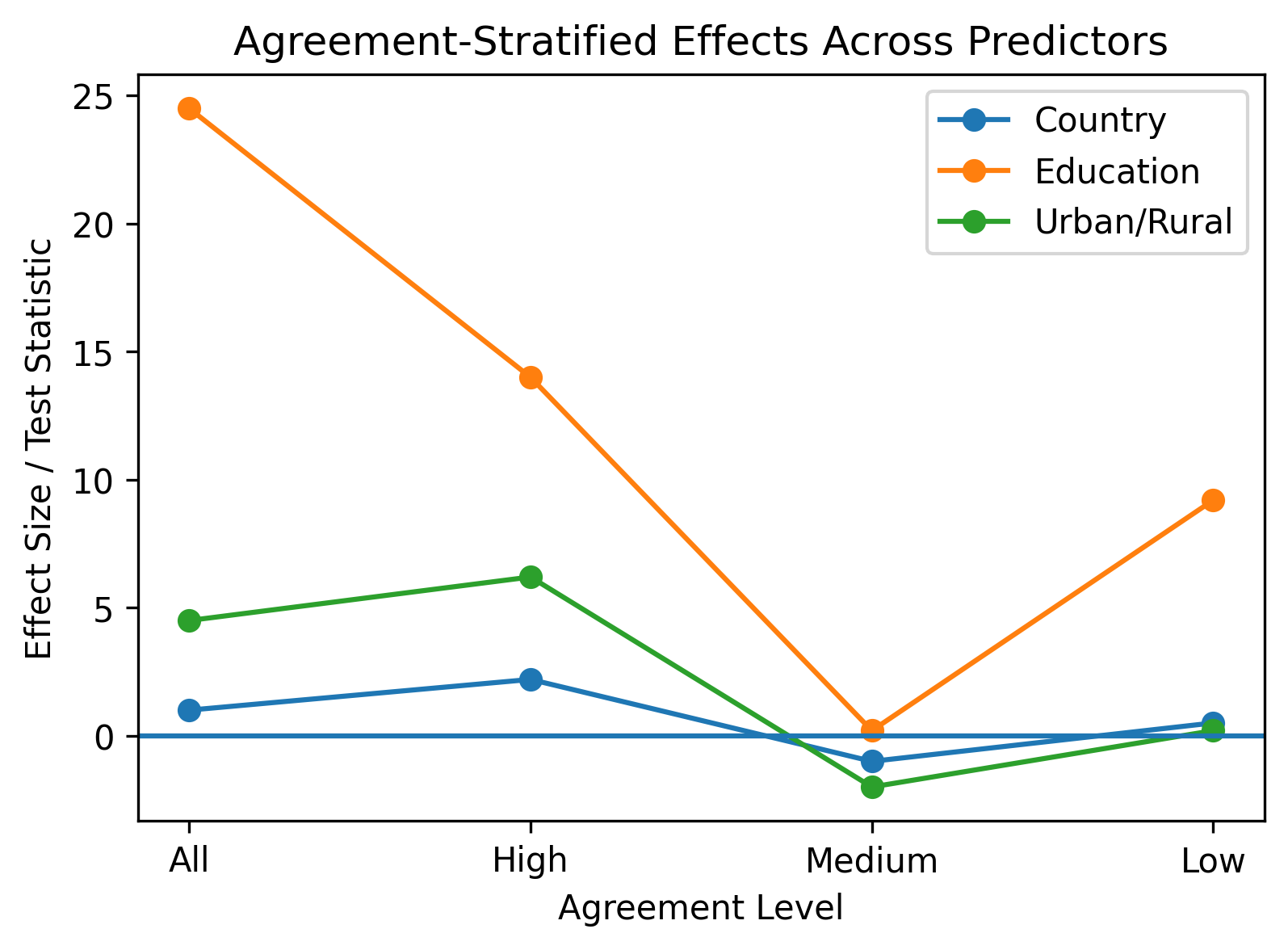}
\caption{Agreement-Stratified Effects Across Predictors
}
\label{fig:agreementall_effects}
\end{figure}

\subsection{Multivariate Interactions}

We further examined whether education effects differ by location and country.
\begin{itemize}
\item \textbf{Location $\times$ Education (All Questions).}
Urban areas showed a significant education effect ($F = 9.7841$, $p < .001$), while rural areas showed an even stronger effect ($F = 12.2732$, $p < .001$). This indicates that education has a stronger association with accuracy in rural areas than in urban areas.

\item \textbf{Country $\times$ Education (All Questions).}
Peru also showed a significant education effect ($F = 3.2749$, $p = .0230$), but the effect was substantially stronger in Ecuador ($F = 14.7899$, $p < .001$). This suggests that the education gradient is more pronounced in Ecuador than in Peru.

\item \textbf{High Agreement Subset.}
In the high-agreement subset, the same general pattern remains visible. The location-by-education effect is still stronger in rural areas (Urban: $F = 2.7109$, $p = .0485$; Rural: $F = 8.8561$, $p < .001$), and the country-by-education effect remains stronger in Ecuador.
\end{itemize}
The multivariate analyses further show that these patterns are context-sensitive: the strength of education effects differs across urban-rural and country comparisons, especially in the high-agreement subset. These multivariate results suggest that education-related differences are not uniform across social contexts, but vary by both geography and country.
\section{Discussion}

This study provides an empirical test of core perspectivist claims within a graded public health evaluation setting. Across analyses, disagreement is shown to be structured rather than incidental.

\paragraph{Structured Disagreement.}
Variance decomposition indicates that disagreement is primarily driven by question-level conceptual difficulty rather than annotator identity. The very low annotator-level ICC suggests that variability does not arise from rater inconsistency, but from properties of the task itself. Disagreement in this setting, therefore, reflects structured epistemic complexity rather than annotation noise.

\paragraph{Disagreement and Inference.}
Agreement-stratified analyses demonstrate that substantive conclusions vary across levels of inter-annotator agreement. Effects that appear robust in aggregated models may strengthen, attenuate, disappear, or reverse direction when evaluated within specific agreement strata. Aggregation thus collapses heterogeneous evaluation contexts into a single estimate, masking instability in underlying relationships.

\paragraph{Beyond Overt Subjectivity.}
These findings extend perspectivist theory beyond clearly subjective domains such as toxicity or stance detection. In a graded health-literacy assessment, disagreement emerges at the intersection of conceptual difficulty and socially distributed knowledge. Variability is not merely interpretive difference, but a signal of uneven epistemic access across respondents.

\paragraph{Methodological Implications.}
The results suggest that strong perspectivist modeling is not simply a normative preference but a requirement for valid inference in graded interpretive tasks. Agreement-sensitive and hierarchical approaches allow researchers to distinguish stable knowledge patterns from contexts of interpretive uncertainty rather than collapsing them into a single summary label.

\paragraph{Practical Implications.}
For NLP, these findings support disagreement-aware evaluation and modeling strategies in tasks that combine factual reasoning with graded interpretation. For public health research, inter-annotator agreement functions as a diagnostic indicator of epistemic stability and should inform how knowledge assessments are interpreted.

Taken together, the results demonstrate that epistemic stability must be modeled explicitly rather than assumed, and that aggregation without regard to disagreement structure risks obscuring substantively important variation.
\section{Conclusion}Health-literacy annotation exhibits structured disagreement driven primarily by conceptual difficulty rather than annotator inconsistency. Substantive social inference varies across levels of inter-annotator agreement: effects that appear robust under aggregation may strengthen, attenuate, or reverse across agreement strata. Agreement-stratified and multivariate analyses further show that these patterns are context-sensitive, with education-related differences varying by geography and country. Together, these results demonstrate that aggregation can obscure meaningful epistemic and social variation, and that disagreement encodes differences in partial knowledge rather than noise. These findings highlight the importance of disagreement-aware approaches for modeling graded, real-world knowledge in NLP and public health settings.

\section{Limitations.}
Several limitations should be noted. First, annotators were drawn from a single institutional context, which may introduce shared interpretive biases despite efforts to standardize training and blind annotations. Second, proportional correctness relies on normative public-health guidelines, which may not fully capture local knowledge practices or alternative valid interpretations. Finally, while agreement-stratified analyses reveal important patterns, they do not establish causal relationships between disagreement and social factors. Future work should develop multilevel and agreement-aware modeling approaches that explicitly incorporate agreement as a moderating factor and extend this framework to other graded evaluation domains.

\section{Ethical Consideratin}
Data were collected as part of a broader health communication study conducted by trained local fieldworkers. Participants provided informed consent for anonymous data collection and analysis. All responses were anonymized prior to annotation and analysis to protect participant privacy. 

\section*{Data Availability}
The deidentified dataset and annotation guidelines are released on GitHub at: \url{https://github.com/olga-kel/Health-Communication}. The resource supports research in low-resource and Indigenous language NLP, multilingual information access, and disagreement-aware annotation modeling.

\section*{Acknowledgments}
The authors thank the local fieldworkers and community collaborators who supported participant recruitment, transcription, and data collection in Ecuador and Peru. In particular, we thank Fernando Ortega, Claudia Crespo, and Marleen Haboud for their contributions to the design and execution of the data collection, as well as their fieldwork assistants for their support in gathering the data. We are especially grateful to Maria Rosa Guamán (Cañar, Ecuador) and Mery Salas Santa Cruz (Apurímac, Peru) for conducting interviews and supporting field data collection. We also thank Talan Herrera and Noah Arellano for translating the datasets from Spanish into English and assisting with annotation. This research was supported by the German Research Foundation (Deutsche Forschungsgemeinschaft, DFG), Project Number 468416293. Olga Kellert was the Principal Investigator (PI) and Stavros Skopeteas was the Co-PI of the funded project that made the data collection possible. Finally, we thank all participants for their time and willingness to take part in the study.
\section*{References}

\bibliographystyle{lrec2026-natbib}
\bibliography{lrec2026}

\end{document}